\documentclass[journal]{IEEEtran}
\usepackage{cite}
\usepackage{amsmath,amssymb,amsfonts}
\usepackage{graphicx}
\usepackage{textcomp}
\usepackage{xcolor}
\usepackage{url} 

\usepackage{array}     
\usepackage{booktabs}  
\usepackage{multirow}
\usepackage{stfloats}  

\usepackage{algorithm}
\usepackage{algorithmic}

\usepackage{newunicodechar}
\newunicodechar{，}{,} 

\newcolumntype{Y}[1]{>{\centering\arraybackslash}p{#1}}
\newcolumntype{M}[1]{>{\centering\arraybackslash}m{#1}}

\begin{document}

\title{InkDiffuser: High-Fidelity One-shot Chinese Calligraphy via Differentiable Morphological Optimization}

\author{Kunchong Shi, Jing Zhang
\thanks{Kunchong Shi and Jing Zhang are with the Department of Computer Science and Engineering, East China University of Science and Technology, Shanghai,200237 China. Corresponding author: Jing Zhang (jingzhang@ecust.edu.cn)}
}
\maketitle

\begin{abstract}

Current Chinese calligraphy generation methods suffer from poor stroke rendering and unrealistic ink morphology, resulting in outputs with limited visual fidelity and artistic fluidity. To address this problem, we propose \textbf{InkDiffuser}, a diffusion-based generative framework for one-shot Chinese calligraphy synthesis. To guarantee high-fidelity rendering, we introduce two core contributions: a high-frequency enhancement mechanism and a Differentiable Ink Structure (DIS) loss that explicitly regularizes ink morphology. Inspired by the observation that high-frequency information in individual samples typically carries contour details, we enhance content extraction by explicitly fusing high-frequency representations for more accurate font structure. Furthermore, we propose a differentiable ink structure loss that integrates differentiable morphological operations into the diffusion process. By allowing the model to learn an explicit decomposition of ink-trace structures, DIS facilitates fine-grained refinement of stroke contours and delivers significantly improved visual realism in the generated calligraphy. Extensive experiments on various calligraphic styles and complex characters demonstrate that InkDiffuser can generate superior calligraphy fonts with realistic ink rendering effects from only a single reference glyph and outperform existing few-shot font generation approaches in structural consistency, detail fidelity, and visual authenticity. The code is available at the following address: \url{https://github.com/JingVIPLab/InkDiffuser}.
\end{abstract}


\IEEEpeerreviewmaketitle

\section{Introduction}

\IEEEPARstart{S}{panning} millennia, the evolution of Chinese characters from oracle bone script to regular script mirrors China's cultural and aesthetic progress, culminating in the revered art form of calligraphy. Chinese calligraphy, as a representative form of visual art, embodies profound cultural heritage and distinctive aesthetic value, still appealing many culture enthusiasts and researchers.

\begin{figure}[t]
    \centering
    \includegraphics[width=1\linewidth]{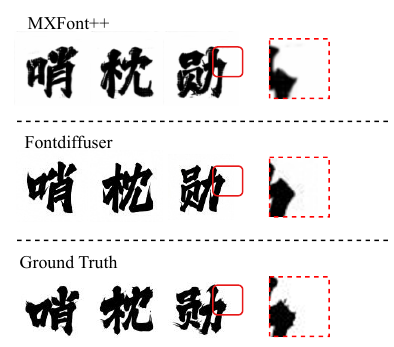}
    \caption{Comparison between real calligraphy (bottom) and results of existing generation methods (top). The red boxes indicate corresponding local regions across all samples to better visualize ink diffusion.}
    \label{fig:motivation}
\end{figure}

In the contemporary era, despite its enduring artistic appeal, mastering calligraphy demands significant time and dedication, which presents a considerable barrier for modern learners. Consequently, the development of font generation techniques, particularly few-shot font generation ~\cite{tang2022few,cha2020few,liu2024deepcallifont}, has emerged as a valuable research direction for preserving and disseminating this art form. 

 Existing few-shot font generation methods can be broadly divided into two paradigms. The first category is GAN-based~\cite{jiang2019scfont,yang2019tet,liao2023calliffusion,liu2024deepcallifont,wen2021handwritten} approaches, which often suffer from high training costs, unstable convergence, and mode collapse. These models also attempt to generate both structural layouts and fine-grained style details in a single step, which proves especially challenging for complex Chinese characters. Consequently, artifacts and blurriness frequently appear around stroke intersections and terminals.

The second category leverages diffusion-based~\cite{ho2020denoising,song2020denoising,song2020score} methods, which incorporate prior knowledge such as stroke sequences or radical compositions to guide the generation process. For instance, Diff-Font~\cite{he2024diff} pioneered the use of diffusion models for font generation by introducing stroke-order priors, significantly improving visual quality. However, for structurally complex characters, encoding and sampling such priors incur substantial computational costs.

Although above methods can synthesize visually pleasing fonts, they primarily generate glyphs that lack the detailed morphological features of real calligraphy, as shown in Fig. \ref{fig:motivation}. Existing methods often yield results that appear overly digital and lack the expressive ink morphology  characteristic of real brush writing, such as MXFont++~\cite{wang2025mx} and Fontdiffuser~\cite{yang2024fontdiffuser}. True brushwork is characterized by intricate and varied ink traces, where ink naturally diffuses and interacts with the paper. While current generative frameworks often oversimplify these visual attributes, producing results that appear flat or blurred rather than faithfully capturing the complex ink morphology. To address this problem, we try to deeply explore how to preserve high-frequency structural details and reproduce realistic ink diffusion dynamics in Chinese calligraphy.

In this paper, we propose InkDiffuser, a diffusion-based image-to-image one-shot Chinese calligraphy generation framework capable of synthesizing unseen calligraphic styles and characters. To improve the morphological accuracy and reproduce complex and realistic ink contours, we introduce high-frequency information and differentiable ink structure loss. High-frequency information, extracted via a dedicated encoder, captures critical structural cues such as stroke edges and sharp corners that are often neglected by standard content encoders. By integrating high-frequency information through our proposed Spatio-Temporal Adaptive Fusion (STAF) module, we can ensure that the generated characters maintain rigorous structural integrity. Inspired by the distinctive visual structure of ink traces, which can be decomposed into a core skeleton and a detailed outer contour, we propose Differentiable Ink Structure (DIS) loss. The key idea of DIS loss is to approximate the non-differentiable \textit{min} and \textit{max} operations in traditional morphological transformations with smooth, differentiable functions, enabling end-to-end gradient optimization. This formulation allows the network to learn such a decomposition, enabling fine-grained control over both the core stroke structure and the precise rendering of the ink contours.

Our main contributions are summarized as follows:
\begin{itemize}
    \item We propose InkDiffuser, a novel diffusion-based image-to-image framework for one-shot brush calligraphy generation,which effectively resolves the challenge of simultaneously preserving intricate structural integrity and rendering realistic ink diffusion dynamics.
    \item To enhance the structural completeness of generated characters, we introduce a spatio-temporal adaptive fusion module that extracts high-frequency components from content images to strengthen stroke representation, thereby improving the integrity of complex character synthesis.        
    \item We design a differentiable ink structure loss  tailored for brush calligraphy. By incorporating soft morphological operators into the optimization process, DIS loss provides explicit supervision on ink trace decomposition, enabling the model to learn realistic ink spread and boundary control in an end-to-end manner.
    \item Extensively experiments on widely used datasets proves that InkDiffuser can achieve the state-of-the-art performance in both structural accuracy and visual ink realism, particularly for complex characters. 

\end{itemize}

\section{Related Work}
\label{sec:related}

In this section, we will introduce some close related works with our method, including: few-shot font generation and one-shot font generation.

\subsection{Few-shot Font Generation}

The core challenge of few-shot font generation~\cite{pan2023few,cha2020few,10483062} lies in constructing a complete, high-quality font library from only a few reference glyphs. This problem is particularly critical for languages with large character sets such as Chinese, where traditional manual font design is time-consuming and costly.

Early font generation approaches relied on stroke extraction and component assembly. With the rise of deep learning—especially Generative Adversarial Networks~\cite{guo2023hgan,wu2020calligan,choi2018stargan,pearton2000gan}, the task gradually evolved into an image-to-image translation problem. Pioneering works such as pix2pix~\cite{isola2017image} demonstrated the potential of GANs for font synthesis; however, they required strictly paired training data, which is scarce in practical font design. To overcome this limitation, models like CycleGAN~\cite{zhu2017unpaired} and StarGAN~\cite{choi2018stargan} introduced cycle-consistency losses to enable style transfer with unpaired datasets, greatly advancing the field. Nevertheless, these general-purpose translation models are not optimized for few-shot settings and still demand multiple reference samples to learn a new style, often resulting in artifacts and mode bias.

In order to address these limitations, researchers have developed specialized few-shot font generation techniques. The mainstream paradigm is content–style disentanglement, which decomposes a glyph image into content features representing structural layout and style features encoding stroke appearance. Early works such as SA-VAE~\cite{sun2017learning} and EMD~\cite{zhang2018separating} employed a variational autoencoder with adversarial learning and a dual-encoder architecture, respectively, to achieve this decomposition, allowing the re-combination of latent content and style representations for new font synthesis. However, simple disentanglement often fails to handle large geometric deformations found in handwriting. To mitigate this, AGIS-Net~\cite{gao2019artistic} introduced a bilinear model to fuse shape and texture features more effectively, while DG-Font~\cite{xie2021dg} incorporated deformable convolutions to align the structural mismatch between source and target fonts unsupervisedly.

Beyond disentanglement, several alternative approaches have been proposed. LF-Font~\cite{park2021few} decomposes characters into radicals for localized stylization, effectively preserving structure but relying heavily on accurate decomposition, limiting its use for cursive calligraphy. MX-Font introduces an external memory module to retrieve relevant stroke patterns from reference samples. Building upon this, XMP-Font~\cite{liu2022xmp} leverages a cross-modality encoder to align glyph images with stroke-level descriptions, reducing the dependency on explicit component labels. CF-Font~\cite{wang2023cf} proposes a content fusion module to bridge the gap between source and target fonts. IF-Font~\cite{chen2024if} replaces source images with Ideographic Description Sequences as structural priors to guide synthesis and VQ-Font~\cite{yao2024vq} introduces a codebook-based prior for structured generation. These approaches provide complementary perspectives beyond pure disentanglement.

\subsection{One-shot Font Generation}

The task of one-shot font generation is inherently ill-posed. It requires a model to extrapolate a global font style from the limited information of a single reference glyph and then render it onto unseen characters with vastly different topologies.
 
General image-to-image translation frameworks like FUNIT~\cite{liu2019few} have been adapted for one-shot font generation by encoding content and style into separate latent spaces. However, although variants of GANs~\cite{azadi2018multi,guo2023hgan,choi2018stargan} have been adapted for this task, their inherent instability and tendency toward mode collapse remain major obstacles. Diffusion models have recently emerged as a powerful alternative, reformulating the generation process as iterative denoising and providing stable training with superior image quality. This paradigm shift has led to significant progress in one-shot font generation and is rapidly becoming the mainstream approach.

Diff-Font~\cite{he2024diff} was the first to introduce diffusion models into font generation, proposing a unified one-shot framework that combines content–style disentanglement with diffusion-based synthesis. By conditioning the denoising process on stroke-order priors, Diff-Font achieves high-fidelity modeling of both stroke morphology and global style. DP-Font~\cite{zhang2024dp} further integrates Physics-Informed Neural Networks into the diffusion framework, utilizing multiple attributes and strict stroke-order constraints to guide calligraphy font generation. More recently, FontDiffuser~\cite{yang2024fontdiffuser} represents a new generation of diffusion-based one-shot font generation methods. It introduces multi-scale content aggregation and style contrastive refinement modules to improve generalization to complex character structures and fine-grained stylistic variations.

\subsection{Summary}

Recent advances in few-shot and one-shot font generation have substantially improved visual quality, yet current methods still fall short for brush calligraphy. They struggle to synthesize complex characters with dense stroke interactions, and often fail to faithfully reproduce the  ink morphology that distinguishes real brushwork from digitally rendered strokes.

To address these challenges, we introduce a novel framework, InkDiffuser, which incorporates explicit structural and morphological priors to generate high-fidelity brush calligraphy. Specifically, high-frequency cues extracted from the content glyph stabilize complex stroke topology, while a differentiable morphological loss supervises the decomposition of ink traces into stroke cores and diffusion boundaries. Extensive experiments proved the effectiveness of InkDuffuser and outperformed state of the art methods.

\section{Background}


Our brush calligraphy font generation framework is built upon the Denoising Diffusion Probabilistic Model~\cite{sasaki2021unit,ho2020denoising} paradigm. This framework consists of a fixed forward noising process and a learned reverse denoising process.

\subsubsection{Forward Process}
The forward process gradually adds Gaussian noise to a clean image $x_0 \sim q(x_0)$ over $T$ timesteps, creating a sequence of latent variables $x_1, \ldots, x_T$. This is modeled as a Markov chain where each step is defined by:

\begin{equation}
    q(x_t|x_{t-1}) = \mathcal{N}(x_t; \sqrt{1-\beta_t}x_{t-1}, \beta_t\mathbf{I}),
    \label{eq:forward_step}
\end{equation}
where $\{\beta_t \in (0, 1)\}_{t=1}^T$ is a predefined variance schedule. A key property of this process is that we can directly sample $x_t$ at any timestep $t$ from the initial state $x_0$ using a closed-form expression:
\begin{equation}
    x_t = \sqrt{\bar{\alpha}_t} x_0 + \sqrt{1 - \bar{\alpha}_t} \epsilon,
    \label{eq:forward_closed_form}
\end{equation}
where $\epsilon \sim \mathcal{N}(0,\mathbf{I})$, $\alpha_t = 1 - \beta_t$, and $\bar{\alpha}_t = \prod_{i=1}^{t} \alpha_i$. As $t \to T$, $x_T$ approaches an isotropic Gaussian distribution.

\subsubsection{Reverse Process}
The reverse process aims to reconstruct a clean sample by starting from pure noise $x_T \sim \mathcal{N}(0,\mathbf{I})$ and iteratively denoising it. This requires approximating the intractable true posterior $p(x_{t-1} | x_t)$.

To achieve this and to guide the generation towards a specific character content and style, we train a conditional neural network $\epsilon_\theta(x_t, t, x_c, x_s)$. Here, $x_c$ is the content condition (e.g., a standard glyph image of the target character) and $x_s$ is the style condition. The network is trained to predict the noise component $\epsilon$ added at timestep $t$.

The model is optimized by minimizing the following simplified objective function, which is a mean squared error between the true and predicted noise:
\begin{equation}
    \mathcal{L} = \mathbb{E}_{t, x_0, \epsilon, x_c, x_s} \left[ \left\| \epsilon - \epsilon_\theta(x_t, t, x_c, x_s) \right\|^2 \right],
    \label{eq:loss_function}
\end{equation}
where $x_t$ is sampled according to Eq.~\eqref{eq:forward_closed_form} and $t$ is uniformly sampled from $\{1, \ldots, T\}$.

\subsubsection{Conditional Sampling}
Once the network $\epsilon_\theta$ is trained, we can generate a new font glyph. Starting from $x_T \sim \mathcal{N}(0,\mathbf{I})$ and given the conditions $x_c$ and $x_s$, we iteratively denoise for $t=T, T-1, \ldots, 1$ using:
\begin{equation}
    x_{t-1} = \frac{1}{\sqrt{\alpha_t}} \left(x_t - \frac{1-\alpha_t}{\sqrt{1 - \bar{\alpha}_t}} \epsilon_\theta(x_t, t, x_c, x_s) \right) + \sigma_t z,
    \label{eq:sampling_step}
\end{equation}
where $z \sim \mathcal{N}(0,\mathbf{I})$ is standard Gaussian noise and the variance $\sigma_t^2$ is a hyperparameter, typically set to $\beta_t$ as proposed in the original work. The final result $x_0$ is the generated glyph image.

\section{Method}
\label{sec:method}

In this section, we will introduce our proposed one-shot Chinese calligraphy generation framework - InkDiffuser in detail.

\begin{figure*}[t]
    \centering
    \includegraphics[width=1\textwidth]{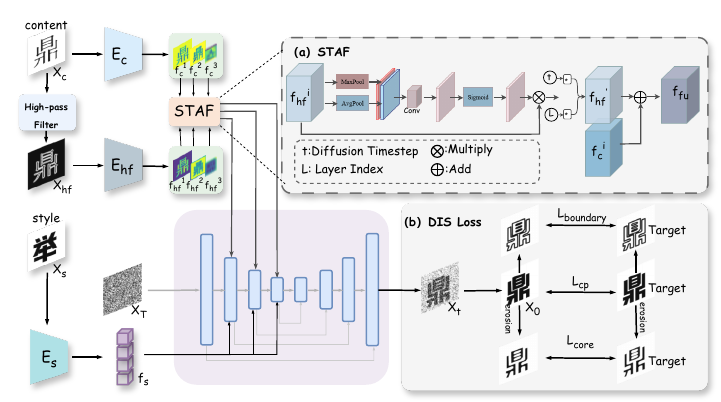}
  \caption{Overview of the proposed InkDiffuser framework. Given a content image $x_c$ and a style image $x_s$, a U-Net denoiser is conditioned on a content encoder $E_c$, a style encoder $E_s$, and a high-frequency encoder $E_{\mathrm{hf}}$. (a) Spatio-Temporal Adaptive Fusion (STAF) module that fuses content and high-frequency features across scales and timesteps. (b) Differentiable Ink Structure (DIS) loss that applies structure- and morphology-aware supervision to the generated ink traces.}
    \label{fig:overview}
\end{figure*}

As shown in Fig.~\ref{fig:overview}, InkDiffuser takes as input a content image $x_c$ and a style image $x_s$. The style image is fed into a style encoder $E_s$ to produce a style embedding $f_s$. In parallel, the content stream is processed through a dual pathway: the content image $x_c$ is passed to a main encoder $E_c$, while a high-frequency version $x_{\mathrm{hf}}$, obtained by applying a high-pass filter to $x_c$, is sent to a dedicated high-frequency encoder $E_{\mathrm{hf}}$. The features from $E_c$ and $E_{\mathrm{hf}}$ are fused by the STAF module to form a unified structural representation $F_c$. Both the style embedding $f_s$ and the structural representation $F_c$ are used as conditions for the main U-Net ~\cite{ronneberger2015u} denoiser $\epsilon_{\theta}$, which reverses the diffusion process and iteratively transforms a noise sample $x_T$ into the final calligraphy image $x_0$. The whole network is optimized with the standard diffusion objective, augmented by the proposed DIS loss that will be detailed in Sec.~\ref{sec:dis_loss}.

\subsection{Spatio-Temporal Adaptive Fusion Module}
High-fidelity generation of complex Chinese characters remains a significant challenge. While existing diffusion-based methods can reconstruct global structures effectively, their content encoders often emphasize low-frequency features~\cite{rahaman2019spectral}, making it difficult to capture stroke edges, corner sharpness, and local textures. This limitation leads to blurry strokes and missing details, particularly in intricate characters.

To address this, we introduce explicit high-frequency modeling. A high-frequency encoder, structurally symmetric to the main content encoder, processes a Sobel-filtered edge map of the content image. This dual-encoder design provides multi-scale content features ($f_c$) and high-frequency features ($f_\text{hf}$) that are perfectly aligned at each layer, forming the basis for our fusion module.

The central challenge lies in fusing these complementary features without introducing artifacts. We propose the STAF module, which formulates this as a purely residual framework. Our method learns to add a hierarchically-modulated detail increment, $f'_{\text{detail}}$, to the content backbone.

This is achieved via a stable, two-level gating mechanism. The final fused feature $f_{\text{fused}}$ is a residual addition, controlled by a global learnable gate $\alpha_{\text{global}}$:
\begin{equation}
    f_{\text{fused}} = f_c + \alpha_{\text{global}} \cdot f'_{\text{detail}},
    \label{eq:global_residual_add}
\end{equation}
where $\alpha_{\text{global}}$ allows the network to learn, in an end-to-end manner, the overall contribution of the high-frequency path, ensuring training stability.

The detail increment $f'_{\text{detail}}$ itself is intelligently crafted by modulating the high-frequency features $f_\text{hf}$ with both spatial and hierarchical adaptive weights:
\begin{equation}
    f'_{\text{detail}} = \alpha(t, l) \cdot (f_{\text{hf}} \odot W_{\text{spatial}}).
    \label{eq:detail_increment}
\end{equation}
This formulation elegantly decouples the fusion into three distinct, interpretable control dimensions: spatial, layer-wise, and temporal.

\subsubsection{Spatial Adaptive Attention}
First, to ensure high-frequency details are applied only where needed, we compute a pixel-wise spatial attention map~\cite{woo2018cbam}. This map identifies salient regions and suppresses noise in smooth areas.
\begin{equation}
    W_{\text{spatial}} = \sigma(\text{Conv}(\text{Align}(f_{\text{hf}}))),
    \label{eq:spatial_attention}
\end{equation}
where $\text{Align}(\cdot)$ ensures channel and spatial alignment with $f_c$.

\subsubsection{Spatio-Temporal Adaptive Weighting}
To achieve dynamic fusion intensity, we introduce a composite weight $\alpha(t, l)$ that modulates the injection of high-frequency details. This weight combines a base learnable parameter $\alpha_{\text{base}}$ with two specific modulation factors:
\begin{equation}
    \alpha(t, l) = \text{clamp}(\alpha_{\text{base}} \cdot \lambda_{\text{layer}}(l) \cdot \lambda_{\text{time}}(t), 0, 1).
    \label{eq:composite_weight}
\end{equation}

The \textit{layer-wise factor} $\lambda_{\text{layer}}(l)$ assigns larger weights to shallow, high-resolution layers to inject fine-grained details, while attenuating the influence in deeper, abstract layers to preserve semantic integrity:
\begin{equation}
    \lambda_{\text{layer}}(l) = \max(0.1, 1.0 - l \cdot \gamma_{\text{layer}}),
    \label{eq:layer_factor}
\end{equation}
where $l$ denotes the layer index and $\gamma_{\text{layer}}$ is set to $0.15$.

Simultaneously, the \textit{time-step factor} $\lambda_{\text{time}}(t)$ provides strong structural guidance during the early, high-noise diffusion stages ($t \approx T$) and gently tapers off as the image clarifies ($t \to 0$) to prevent artifacts:
\begin{equation}
    \lambda_{\text{time}}(t) = 1.0 + \frac{t}{T} \cdot \gamma_{\text{time}},
    \label{eq:time_factor}
\end{equation}
where $\gamma_{\text{time}}$ is empirically set to $0.2$.

In summary, STAF implements a highly efficient and stable fusion strategy. By decomposing the fusion into a pure residual addition of a three-dimensionally controlled detail increment (spatial, layer-wise, and temporal), our method dramatically improves the structural integrity and detail fidelity of complex characters with minimal computational overhead.

\subsection{Differentiable Ink Structure Loss}
\label{sec:dis_loss}

Generating calligraphy that captures desirable visual aesthetics, particularly in rendering the morphological details of ink contours and their transition zones, remains a significant challenge. Traditional pixel-wise loss functions such as L1 or MSE~\cite{chicco2021coefficient} treat the image as a monolithic entity, failing to distinguish between the sharp core of a stroke and the delicate, often blurred, ink diffusion at its edges. This limitation makes it exceedingly difficult to concurrently preserve the structural integrity of the strokes while precisely optimizing the geometric properties of the ink contour.
 
 To address this core challenge, we propose the Differentiable Ink Structure (DIS) loss. The key idea of the DIS loss is Decoupled Control, which decomposes the complex ink generation problem into three independent yet complementary objectives: preserve the structural consistency of stroke cores, precisely control the extent of the diffusion boundaries and ensure smooth and natural diffusion edges.

These three objectives are jointly enforced by three specifically designed loss components. In traditional morphological operations, such as erosion and dilation~\cite{gil2003efficient}, the operations are typically implemented by taking the minimum or maximum value within a local neighborhood. However, the min and max functions are non-differentiable. When incorporated into neural networks, gradients cannot propagate through them. Consequently, any loss function relying on conventional morphological operations cannot be used for end-to-end deep learning, severely limiting the use of geometric information to optimize the model.

\begin{figure}[h] %
    \centering
    \includegraphics[width=0.85\linewidth]{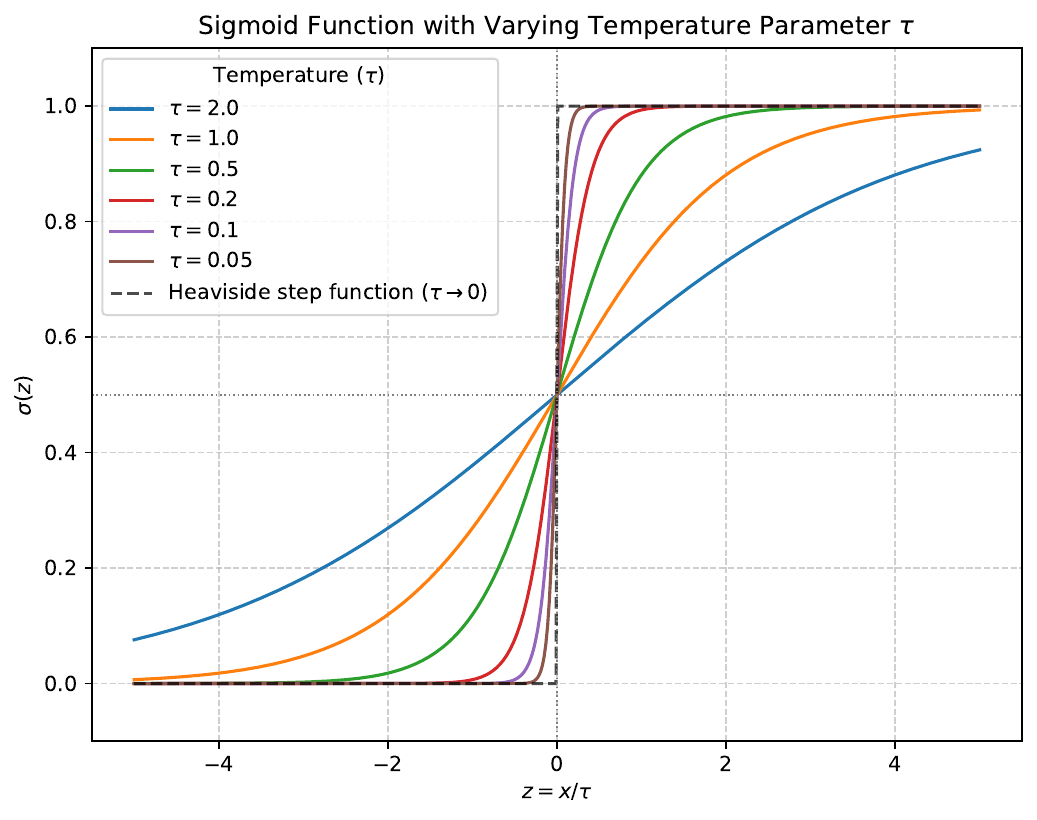} 
    \caption{Illustration of the sigmoid function $\sigma(z/\tau)$ with varying temperature parameters $\tau$. As $\tau$ decreases, the curve approaches a sharp step function, allowing for a differentiable approximation of the hard logic operators.}
    \label{fig:sigmoid_tau_effect}
\end{figure}

To overcome this, we formulate ``soft'' differentiable versions of these operations. The key insight is to approximate the non-differentiable $\min$/$\max$ operators with a smooth function whose behavior can be precisely controlled. We achieve this using the sigmoid~\cite{han1995influence} function, scaled by a temperature parameter $\tau > 0$. The crucial property of this formulation lies in its limit behavior: as the temperature $\tau$ approaches zero, the function $\sigma(z/\tau)$ converges to a Heaviside step function, as shown in Fig. \ref{fig:sigmoid_tau_effect}. This mimics the ``hard'', non-differentiable nature of the original operators. However, for a small but positive $\tau$, the transition is smooth and continuously differentiable, allowing gradients to flow during backpropagation. The parameter $\tau$ thus controls the ``softness'' of the approximation.

In our DIS Loss, we implement these soft morphological operations. The convolution operation, $\text{conv}(x, \text{kernel})$, aggregates information from the neighborhood defined by the circular structuring element ($\text{kernel}$), analogous to the role of the structuring element in classical morphology. The soft erosion, which approximates the $\min$ function, is formulated as:
\begin{equation}
    \text{soft\_erosion}(x) = -\sigma(-\text{conv}(x, \text{kernel})/\tau) \cdot \tau.
    \label{eq:soft_erosion}
\end{equation}
Conversely, the soft dilation, which approximates the $\max$ function, is formulated as:
\begin{equation}
    \text{soft\_dilation}(x) = \sigma(\text{conv}(x, \text{kernel})/\tau) \cdot \tau.
    \label{eq:soft_dilation}
\end{equation}
This differentiable formulation enables the entire DIS Loss, including its morphological analysis components, to be integrated seamlessly into our end-to-end training pipeline.

In calligraphy, both the transitions of strokes and the desired ink spread exhibit smooth and rounded geometric characteristics. To prevent the "stair-step" artifacts introduced by traditional square structuring elements on calligraphic curves, we employ circular structuring elements to better preserve the natural smoothness of stroke contours during morphological operations.

\subsubsection{Overall Architecture of DIS Loss}

The DIS Loss is formulated as a weighted sum of its three core components:
\begin{equation}
L_\text{DIS} = \lambda_\text{c} L_\text{c} + \lambda_\text{b} L_\text{b} + \lambda_\text{lap} L_\text{lap},
\end{equation}
where $L_\text{c}$ is the core structural consistency loss, $L_\text{b}$ is the diffusion boundary consistency loss, and $L_\text{lap}$ is the diffusion smoothness loss. The hyperparameters $\lambda_\text{c}, \lambda_\text{b}, \lambda_\text{lap}$ control the relative importance of each component during training.

\subsubsection{Core Structural Consistency Loss}  
This loss ensures the skeletal structure of generated fonts remains accurate, even under strong ink diffusion effects. To focus solely on the stroke cores, we apply the differentiable soft erosion operation to strip away edge pixels, isolating the core regions of the strokes. $L_\text{c}$ is then computed as the L1 distance between the eroded structures of the generated image $I_\text{generated}$ and the target image $I_\text{target}$:
\begin{equation}
L_\text{c} = \| \text{Erode}(I_\text{generated}) - \text{Erode}(I_\text{target}) \|_1.
\end{equation}
Unlike traditional erosion, our soft erosion is differentiable, allowing effective gradient propagation during training.

\subsubsection{Diffusion Boundary Consistency Loss}  
To precisely control the extent and shape of ink diffusion, we design the diffusion boundary consistency loss $L_\text{b}$. We first generate a boundary mask $M_\text{boundary}$ by computing the difference between the dilated and eroded target image:
\begin{equation}
M_\text{boundary} = \text{Dilate}(I_\text{target}) - \text{Erode}(I_\text{target}).
\end{equation}
This mask isolates the critical diffusion regions, allowing the loss to focus only on relevant ink-spread areas. The loss is then calculated as:
\begin{equation}
L_\text{b} = \| (I_\text{generated} - I_\text{target}) \odot M_\text{boundary} \|_1.
\end{equation}

\subsubsection{Diffusion Smoothness Loss}  
The edges of real ink diffusion exhibit a smooth transition from blurred to sharp. To encourage this property, the diffusion smoothness loss $L_\text{lap}$ employs the Laplacian~\cite{merris1994laplacian} operator to detect edge responses. By minimizing the difference between the Laplacian-filtered generated and target images, we effectively constrain the smoothness of stroke edges:
\begin{equation}
L_\text{lap} = \| \text{Laplacian}(I_\text{generated}) - \text{Laplacian}(I_\text{target}) \|_1.
\end{equation}

Through the synergistic effect of these three components, the DIS Loss guides the model to generate fonts that maintain structural accuracy while precisely controlling and reproducing the highly realistic and aesthetically pleasing morphology of ink contours.

\subsection{Training Objective}

During training, we primarily adopt the standard MSE diffusion loss to optimize InkDiffuser, ensuring the generator’s capability for faithful font reconstruction:
\begin{equation}
\mathcal{L}_{\text{total}}
  = \mathcal{L}_{\text{MSE}}
  + \lambda_{\text{cp}}\,\mathcal{L}_{\text{cp}}
  + \lambda_{\text{dis}}\,\mathcal{L}_{\text{dis}},
\end{equation}

\begin{equation}
\mathcal{L}_{\text{MSE}} = \left\lVert \epsilon - \epsilon_{\theta}(x_t, t, x_c, x_s) \right\rVert_2^{2},
\end{equation}
\begin{equation}
\mathcal{L}_{\text{cp}} = \sum_{l=1}^{L} \left\lVert {VGG}_l(x_{0}) - {VGG}_l(x_{\text{target}}) \right\rVert_{1},
\end{equation}

\begin{equation}
\mathcal{L}_{\text{dis}} = \lambda_{\text{c}}\,\mathcal{L}_{\text{c}}
 + \lambda_{\text{b}}\,\mathcal{L}_{\text{b}}
 + \lambda_{\text{lap}}\,\mathcal{L}_{\text{lap}}.
\end{equation}

Here, $L_\text{total}$ denotes the overall training loss. $\text{VGG}_l(\cdot)$ represents features encoded by the $l$-th layer of a pre-trained VGG network, and $L$ is the number of selected layers. $L_\text{cp}$ penalizes misalignment between the VGG features of the generated $x_0$ and the corresponding target $x_\text{target}$. Through extensive experiments, the hyperparameters are set as $\lambda_\text{cp} = 0.01$, $\lambda_\text{dis} = 0.02$.

\section{Experiments }
\label{sec:exp_results}

This section provides a thorough evaluation of InkDiffuser, covering comparative results, ablation studies, hyperparameter analyses, and an assessment of its limitations.

\subsection{Datasets and Evaluation Metrics}
\label{subsec:datasets}

Our dataset comprises 293 high-quality Chinese fonts collected from the Foundertype library~\cite{foundertype}, split into 269 seen fonts for training and 24 unseen fonts for testing. The training set utilizes 1000 commonly used characters to ensure consistent coverage. For evaluation, we define two settings: UFUC (13 unseen fonts generated on 212 unseen characters) and UFSC (11 unseen fonts generated on 274 seen characters). This setup allows us to assess the model's generalization across both novel styles and novel content.

Five widely used metrics are used to evaluate our model performance, including RMSE, PSNR, SSIM~\cite{wang2004image}, LPIPS~\cite{zhang2018unreasonable} and L1 loss. These metrics comprehensively reflect the visual quality, structural fidelity, and perceptual similarity of generated characters.

\subsection{Implementation Details}
\label{subsec:implementation}

We train InkDiffuser using the AdamW optimizer with parameters $\beta_1=0.9$ and $\beta_2=0.999$. The model is trained for 440,000 steps with a batch size of 16. The learning rate is initialized at $1\times10^{-4}$ with a linear schedule. To ensure consistent character scale, all characters are resized to $96 \times 96$ pixels, which helps eliminate size-related variance and facilitates fair comparison across methods. All experiments are conducted on a single NVIDIA RTX 3090 GPU.

\subsection{Comparison Experiments}
\label{subsec:sota_comparison}

We compare InkDiffuser with five representative approaches: CF-Font~\cite{wang2023cf}, IF-Font~\cite{chen2024if}, MSD-Font~\cite{fu2024generate}, FontDiffuser~\cite{yang2024fontdiffuser} and MXFont++~\cite{wang2025mx}. These baselines cover diverse paradigms, including GAN-based, component-based, and diffusion-based methods, providing a holistic benchmark for performance evaluation.

For a fair comparison, we retrained all baseline models on the same dataset under identical training conditions.

\begingroup
\renewcommand{\arraystretch}{1.18} 
\setlength{\extrarowheight}{0.2ex} 
\begin{table*}[t!]
\centering 
\caption{Quantitative comparison results on UFUC and UFSC datasets.}
\label{tab:metric}
\begin{tabular*}{0.8\textwidth}{@{\extracolsep{\fill}}c|c|ccccc}
\hline
Datasets & Methods & LPIPS$\downarrow$ & SSIM$\uparrow$ & L1$\downarrow$ & RMSE$\downarrow$ & PSNR$\uparrow$ \\ \hline
\multirow{6}{*}{UFUC} & CF-Font~\cite{wang2023cf} & 0.1673 & 0.5682 & 0.1764 & 0.3392 & 9.7721 \\
 & IF-Font~\cite{chen2024if} & 0.1685 & 0.4534 & 0.1860 & 0.4117 & 9.7534 \\
 & MSD-Font~\cite{fu2024generate} & 0.1787 & 0.5646 & 0.1447 & 0.3275 & 9.7461 \\
 & Fontdiffuser~\cite{yang2024fontdiffuser} & 0.1615 & 0.5857 & 0.1309 & 0.3241 & 9.8370 \\
 & MXFont++~\cite{wang2025mx} & 0.2038 & 0.5531 & 0.1432 & 0.3239 & 9.8475 \\
 & InkDiffuser & \textbf{0.1573} & \textbf{0.6099} & \textbf{0.1237} & \textbf{0.3150} & \textbf{10.0861} \\ \hline \hline
\multirow{6}{*}{UFSC} & CF-Font~\cite{wang2023cf} & 0.1485 & 0.5837 & 0.1196 & 0.3102 & 10.0823 \\
 & IF-Font~\cite{chen2024if} & 0.1573 & 0.5417 & 0.1532 & 0.3728 & 9.867 \\
 & MSD-Font~\cite{fu2024generate} & 0.1489 & 0.5868 & 0.1324 & 0.3081 & 10.3407 \\
 & Fontdiffuser~\cite{yang2024fontdiffuser} & 0.1416 & 0.5998 & 0.1190 & 0.3050 & 10.4069 \\
 & MXFont++~\cite{wang2025mx} & 0.1952 & 0.5550 & 0.1370 & 0.3143 & 10.1473 \\
 & InkDiffuser & \textbf{0.1414} & \textbf{0.6087} & \textbf{0.1163} & \textbf{0.3017} & \textbf{10.4785} \\ \hline
\end{tabular*}
\end{table*}
\endgroup

\subsubsection{Quantitative Comparison}
\label{subsubsec:quantitative}

Table~\ref{tab:metric} presents the quantitative comparison results against other methods. Overall, InkDiffuser consistently achieves the best performance across all five metrics on both the UFUC and UFSC datasets, demonstrating its superiority in both structural fidelity and perceptual quality.

The UFUC benchmark represents the most challenging scenario, requiring the model to generalize to unseen character structures. 
As shown in Table~\ref{tab:metric}, our method exhibits significant improvements over the strongest competitor, FontDiffuser. 
Specifically, InkDiffuser boosts the structural similarity metric (SSIM) by a margin of 0.0242 and reduces the pixel-level L1 error by 0.0072. 
Notably, in terms of signal-to-noise ratio, we achieve a PSNR gain of approximately 0.25 dB compared to the runner-up.
These substantial margins in structural metrics (SSIM and L1) validate the effectiveness of our STAF module, which successfully injects high-frequency cues to prevent the structural collapse often observed in baseline methods when handling unseen glyphs.

In the UFSC setting, where character structures are known but styles are unseen, InkDiffuser maintains its leadership. 
Compared to FontDiffuser, our method further lowers the perceptual distance (LPIPS) and decreases the Root Mean Square Error (RMSE) by 0.0033. 
Although the baseline methods already perform reasonably well in this easier setting, InkDiffuser still squeezes out performance gains, particularly in reproducing fine-grained ink details. 
This consistent superiority across both benchmarks confirms that the proposed DIS loss effectively guides the model to synthesize realistic ink morphology, regardless of whether the character topology is seen or unseen.

\subsubsection{Qualitative Comparison}
\label{subsubsec:qualitative}

\begin{figure*}[t!]
    \centering
    \includegraphics[width=1\textwidth]{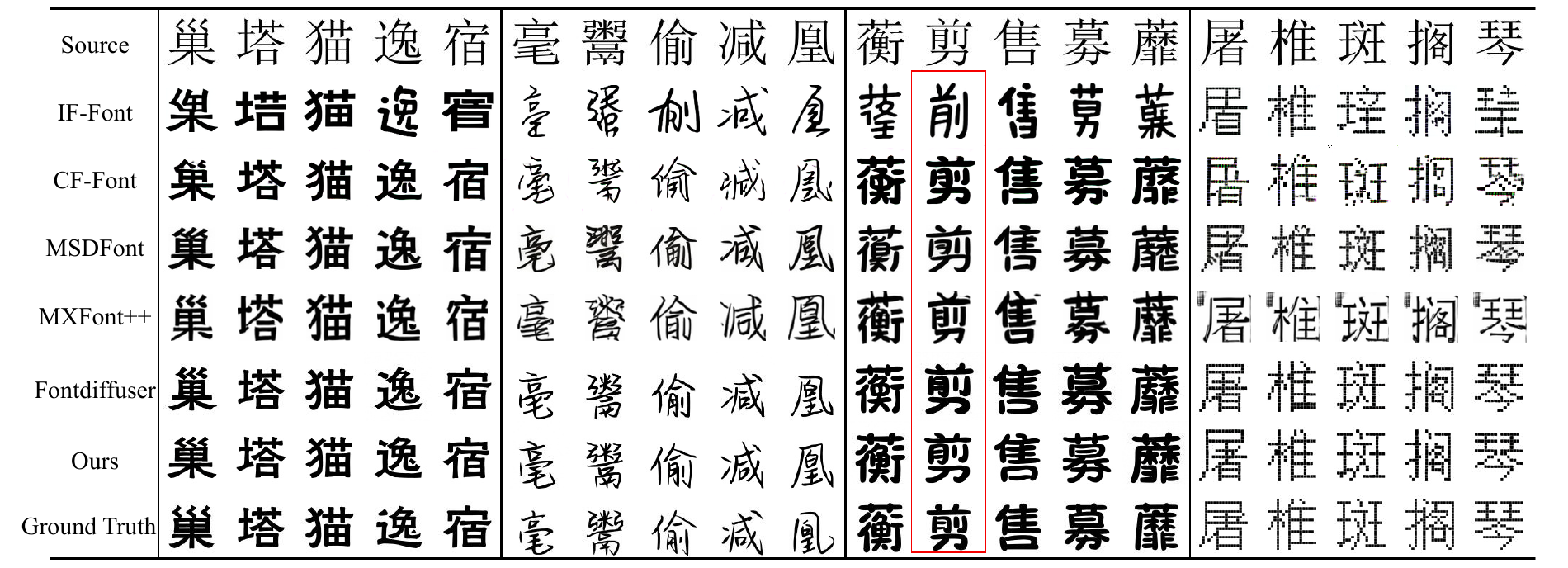}
    \caption{Qualitative comparison on the UFUC setting. The results generated by InkDiffuser are closest to the target fonts and contain noticeably fewer wrongly written characters.}

    \label{fig:ufuc_visible}
\end{figure*}

The qualitative results are shown in Fig.~\ref{fig:ufuc_visible}, where we select four fonts, including two simple fonts and two complex fonts along with 20 Chinese characters. All characters are generated by six font generation models for comparison. Consider the character ``\textit{Jian}'' which is displayed in the 12th column as an example. IF-Font~\cite{chen2024if} produces stroke errors and struggles to maintain the structural integrity of character. CF-Font~\cite{wang2023cf} and MXFont++~\cite{wang2025mx} suffer from missing strokes and produce blurred glyphs. MSD-Font~\cite{fu2024generate} preserves the overall character shapes, but the generated character lacks clarity and sharpness. FontDiffuser~\cite{yang2024fontdiffuser} demonstrates decent performance, producing structurally coherent characters without obvious errors. However, a noticeable gap remains compared to the ground truth, as it struggles to fully capture the authentic brushwork fidelity. In contrast, InkDiffuser consistently outperforms all other methods across this case, demonstrating its effectiveness in generating high-quality Chinese characters with accurate structural details and clear stroke rendering.

\subsection{Ablation Study}
\label{subsec:ablation}

In this section, we conduct an ablation study to analyze the contributions of the proposed STAF module and DIS loss. All experiments are evaluated on the Unseen Fonts and Unseen Contents test set.

\begin{table}[h]
\renewcommand{\arraystretch}{1.18} 
\setlength{\extrarowheight}{0.2ex} 
\centering
\small 
\setlength{\tabcolsep}{4pt} 
\caption{Ablation study on the effectiveness of the STAF and DIS modules.}
\label{tab:ablation_single_col_sep}
\begin{tabular*}{\columnwidth}{@{\extracolsep{\fill}}ccccccc}
\toprule
\multicolumn{2}{c}{Module} & \multirow{2}{*}{LPIPS$\downarrow$} & \multirow{2}{*}{SSIM$\uparrow$} & \multirow{2}{*}{L1$\downarrow$} & \multirow{2}{*}{RMSE$\downarrow$} & \multirow{2}{*}{PSNR$\uparrow$} \\
\cmidrule(r){1-2}
STAF & DIS & & & & & \\
\midrule
$\times$ & $\times$ & 0.1615 & 0.5857 & 0.1309 & 0.3241 & 9.8370 \\
$\checkmark$ & $\times$ & 0.1675 & 0.5973 & 0.1288 & 0.3222 & 9.8894 \\
$\times$ & $\checkmark$ & 0.1596 & 0.5945 & 0.1285 & 0.3210 & 9.9215 \\
$\checkmark$ & $\checkmark$ & \textbf{0.1573} & \textbf{0.6099} & \textbf{0.1237} & \textbf{0.3150} & \textbf{10.0861} \\
\bottomrule
\end{tabular*}
\end{table}

\begin{figure}[t]
    \centering
    \includegraphics[width=1\linewidth]{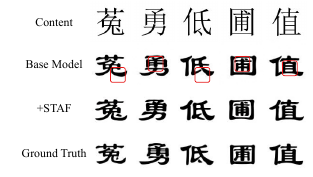}
    \caption{Qualitative results of the ablation study on the STAF module.}
    \label{fig:STAF}
\end{figure}

\begin{figure}[t]
    \centering
    \includegraphics[width=1\linewidth]{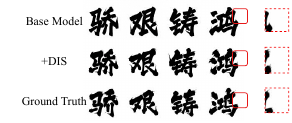}
    \caption{Qualitative results of the ablation study on the DIS module.}
    \label{fig:DIS}
\end{figure}

The baseline model, a standard conditional diffusion model, often suffers from structural collapse. As illustrated in Fig.~\ref{fig:STAF}, dense strokes in complex characters are prone to disappearing or blurring. This limitation stems from the inherent spectral bias of deep neural networks, which tend to prioritize low-frequency global shapes over high-frequency local details. 

By incorporating the STAF module, we explicitly inject high-frequency priors into the denoising process. Quantitative results show a clear improvement in structural metrics, with SSIM rising significantly and L1 error decreasing. This indicates that STAF effectively compensates for information loss, restoring the missing topological details. Interestingly, we observe a slight increase in LPIPS when using STAF alone. This is because STAF sharpens the character edges to ensure structural correctness, but without the morphological supervision of DIS, these high-frequency edges may appear overly sharp compared to the smooth ink flow of the ground truth, leading to a penalty in perceptual metrics.

The DIS loss targets the physical properties of brush handwriting. Adding DIS alone to the baseline improves all metrics, particularly LPIPS, which drops below the baseline level. Visual results in Fig.~\ref{fig:DIS} confirm that DIS helps to synthesize realistic ink seepage and diffusion boundaries, making the generated glyphs visually more authentic.

The full model integrates both modules to achieve the best performance. The combination of STAF's structural guidance and DIS's morphological regularization creates a synergistic effect. The sharp edges produced by STAF are softened and naturalized by DIS, resulting in the lowest LPIPS score and the highest SSIM.

\subsection{Experiments on Hyperparameter $\lambda_{\mathrm{dis}}$}    
\label{subsec:hyperparameter}

The hyperparameter $\lambda_{\mathrm{dis}}$ governs the strength of morphological supervision, acting as a trade-off knob between ink realism and structural preservation. We investigated the sensitivity of our model to this weight by varying $\lambda_{\mathrm{dis}}$ from 0 to 0.04 on the UFUC dataset, with results summarized in Table~\ref{tab:ablation_lambda}.

Introducing the DIS loss yields immediate performance gains. As $\lambda_{\mathrm{dis}}$ increases to 0.02, we observe a monotonic improvement across all structural and pixel-level metrics. Notably, compared to the baseline (w/o DIS), the model at $\lambda_{\mathrm{dis}} = 0.02$ achieves a substantial boost in SSIM and a significant reduction in L1 error. This confirms that the proposed loss successfully guides the diffusion process to refine ink boundaries and stroke transitions without interfering with the global glyph topology.

The performance peaks at $\lambda_{\mathrm{dis}} = 0.02$. While increasing the weight further to 0.03 results in a negligible drop in perceptual distance (LPIPS), it causes a sharp decline in structural fidelity (SSIM). This phenomenon suggests that excessive weight on ink morphology forces the model to hallucinate realistic-looking textures that may not align with the ground truth structure, effectively prioritizing style over content. Pushing $\lambda_{\mathrm{dis}}$ to 0.04 exacerbates this issue, causing all metrics to deteriorate below the optimal levels. Therefore, we fix $\lambda_{\mathrm{dis}} = 0.02$ as the robust setting to ensure high-fidelity generation in both structure and texture.

\begin{table}[h]
\centering
\caption{Sensitivity analysis of the DIS loss weight $\lambda_{\mathrm{dis}}$ on the UFUC dataset. The optimal setting is highlighted in bold.}
\label{tab:ablation_lambda}
\renewcommand{\arraystretch}{1.15}
\setlength{\tabcolsep}{3.5pt}
\begin{tabular}{lccccc}
\toprule
$\lambda_{\mathrm{dis}}$ & LPIPS$\downarrow$ & SSIM$\uparrow$ & L1$\downarrow$ & RMSE$\downarrow$ & PSNR$\uparrow$ \\
\midrule
0 (w/o DIS) & 0.1596 & 0.5945 & 0.1285 & 0.3210 & 9.9215 \\
0.01        & 0.1583 & 0.5967 & 0.1275 & 0.3196 & 9.9812 \\
0.02        & 0.1573 & \textbf{0.6099} & \textbf{0.1237} & \textbf{0.3150} & \textbf{10.0861} \\
0.03        & \textbf{0.1569} & 0.5975 & 0.1262 & 0.3165 & 10.0458 \\
0.04        & 0.1582 & 0.5912 & 0.1269 & 0.3214 & 9.9338 \\
\bottomrule
\end{tabular}
\end{table}

\section{Conclusion}
\label{sec:conclusion}

We presented InkDiffuser, a diffusion-based one-shot Chinese calligraphy generator equipped with a spatio-temporal adaptive fusion module and a differentiable ink structure loss. By injecting high-frequency structural cues and differentiable morphological priors into a conditional diffusion model, InkDiffuser substantially improves structural fidelity and ink realism over prior methods. Extensive experiments on challenging Chinese calligraphy datasets demonstrate that our method can achieve state-of-the-art performance, outperforming existing baselines in both quantitative metrics and visual quality. In future work, we plan to tackle the challenge of generating highly abstract artistic styles by developing adaptive structural priors.

\ifCLASSOPTIONcaptionsoff
  \newpage
\fi

\bibliographystyle{IEEEtran}

\bibliography{reference}

@String(ECCV= {Eur. Conf. Comput. Vis.})

@String(TOG= {ACM Trans. Graph.})

@String(ICASSP=	{ICASSP})

@String(AAAI = {AAAI})

@String(ECCV  = {ECCV})

@String(TOG   = {ACM TOG})

@inproceedings{azadi2018multi,
  title={Multi-content gan for few-shot font style transfer},
  author={Azadi, Samaneh and Fisher, Matthew and Kim, Vladimir G and Wang, Zhaowen and Shechtman, Eli and Darrell, Trevor},
  booktitle={Proceedings of the IEEE conference on computer vision and pattern recognition},
  pages={7564--7573},
  year={2018}
}

@inproceedings{liu2024deepcallifont,
  title={DeepCalliFont: Few-shot Chinese calligraphy font synthesis by integrating dual-modality generative models},
  author={Liu, Yitian and Lian, Zhouhui},
  booktitle={Proceedings of the AAAI conference on artificial intelligence},
  volume={38},
  number={4},
  pages={3774--3782},
  year={2024}
}

@inproceedings{ronneberger2015u,
  title={U-net: Convolutional networks for biomedical image segmentation},
  author={Ronneberger, Olaf and Fischer, Philipp and Brox, Thomas},
  booktitle={International Conference on Medical image computing and computer-assisted intervention},
  pages={234--241},
  year={2015},
  organization={Springer}
}

@article{wang2004image,
  title={Image quality assessment: from error visibility to structural similarity},
  author={Wang, Zhou and Bovik, Alan C and Sheikh, Hamid R and Simoncelli, Eero P},
  journal={IEEE transactions on image processing},
  volume={13},
  number={4},
  pages={600--612},
  year={2004},
  publisher={IEEE}
}

@inproceedings{zhang2018unreasonable,
  title={The unreasonable effectiveness of deep features as a perceptual metric},
  author={Zhang, Richard and Isola, Phillip and Efros, Alexei A and Shechtman, Eli and Wang, Oliver},
  booktitle={Proceedings of the IEEE conference on computer vision and pattern recognition},
  pages={586--595},
  year={2018}
}

@inproceedings{woo2018cbam,
  title={Cbam: Convolutional block attention module},
  author={Woo, Sanghyun and Park, Jongchan and Lee, Joon-Young and Kweon, In So},
  booktitle={Proceedings of the European conference on computer vision (ECCV)},
  pages={3--19},
  year={2018}
}

@inproceedings{rahaman2019spectral,
  title={On the spectral bias of neural networks},
  author={Rahaman, Nasim and Baratin, Aristide and Arpit, Devansh and Draxler, Felix and Lin, Min and Hamprecht, Fred and Bengio, Yoshua and Courville, Aaron},
  booktitle={International conference on machine learning},
  pages={5301--5310},
  year={2019},
  organization={PMLR}
}

@inproceedings{cha2020few,
  title={Few-shot compositional font generation with dual memory},
  author={Cha, Junbum and Chun, Sanghyuk and Lee, Gayoung and Lee, Bado and Kim, Seonghyeon and Lee, Hwalsuk},
  booktitle={European conference on computer vision},
  pages={735--751},
  year={2020},
  organization={Springer}
}

@article{gao2019artistic,
  title={Artistic glyph image synthesis via one-stage few-shot learning},
  author={Gao, Yue and Guo, Yuan and Lian, Zhouhui and Tang, Yingmin and Xiao, Jianguo},
  journal={ACM Transactions on Graphics (ToG)},
  volume={38},
  number={6},
  pages={1--12},
  year={2019},
  publisher={ACM New York, NY, USA}
}

@inproceedings{jiang2019scfont,
  title={Scfont: Structure-guided chinese font generation via deep stacked networks},
  author={Jiang, Yue and Lian, Zhouhui and Tang, Yingmin and Xiao, Jianguo},
  booktitle={Proceedings of the AAAI conference on artificial intelligence},
  volume={33},
  number={01},
  pages={4015--4022},
  year={2019}
}

@article{guo2023hgan,
  title={HGAN: Hierarchical graph alignment network for image-text retrieval},
  author={Guo, Jie and Wang, Meiting and Zhou, Yan and Song, Bin and Chi, Yuhao and Fan, Wei and Chang, Jianglong},
  journal={IEEE Transactions on Multimedia},
  volume={25},
  pages={9189--9202},
  year={2023},
  publisher={IEEE}
}

@inproceedings{yang2019tet,
  title={TET-GAN: Text effects transfer via stylization and destylization},
  author={Yang, Shuai and Liu, Jiaying and Wang, Wenjing and Guo, Zongming},
  booktitle={Proceedings of the AAAI Conference on Artificial Intelligence},
  volume={33},
  number={01},
  pages={1238--1245},
  year={2019}
}

@inproceedings{pan2023few,
  title={Few shot font generation via transferring similarity guided global style and quantization local style},
  author={Pan, Wei and Zhu, Anna and Zhou, Xinyu and Iwana, Brian Kenji and Li, Shilin},
  booktitle={Proceedings of the IEEE/CVF International Conference on Computer Vision},
  pages={19506--19516},
  year={2023}
}

@ARTICLE{10483062,
  author={He, Xiao and Zhu, Mingrui and Wang, Nannan and Gao, Xinbo},
  journal={IEEE Transactions on Circuits and Systems for Video Technology}, 
  title={Few-Shot Font Generation by Learning Style Difference and Similarity}, 
  year={2024},
  volume={34},
  number={9},
  pages={8013-8025},
  doi={10.1109/TCSVT.2024.3382621}}

@article{liao2023calliffusion,
  title={Calliffusion: Chinese calligraphy generation and style transfer with diffusion modeling},
  author={Liao, Qisheng and Xia, Gus and Wang, Zhinuo},
  journal={arXiv preprint arXiv:2305.19124},
  year={2023}
}

@article{he2024diff,
  title={Diff-font: Diffusion model for robust one-shot font generation},
  author={He, Haibin and Chen, Xinyuan and Wang, Chaoyue and Liu, Juhua and Du, Bo and Tao, Dacheng and Yu, Qiao},
  journal={International Journal of Computer Vision},
  volume={132},
  number={11},
  pages={5372--5386},
  year={2024},
  publisher={Springer}
}

@inproceedings{isola2017image,
  title={Image-to-image translation with conditional adversarial networks},
  author={Isola, Phillip and Zhu, Jun-Yan and Zhou, Tinghui and Efros, Alexei A},
  booktitle={Proceedings of the IEEE conference on computer vision and pattern recognition},
  pages={1125--1134},
  year={2017}
}

@article{ho2020denoising,
  title={Denoising diffusion probabilistic models},
  author={Ho, Jonathan and Jain, Ajay and Abbeel, Pieter},
  journal={Advances in neural information processing systems},
  volume={33},
  pages={6840--6851},
  year={2020}
}

@article{song2020denoising,
  title={Denoising diffusion implicit models},
  author={Song, Jiaming and Meng, Chenlin and Ermon, Stefano},
  journal={arXiv preprint arXiv:2010.02502},
  year={2020}
}

@article{song2020score,
  title={Score-based generative modeling through stochastic differential equations},
  author={Song, Yang and Sohl-Dickstein, Jascha and Kingma, Diederik P and Kumar, Abhishek and Ermon, Stefano and Poole, Ben},
  journal={arXiv preprint arXiv:2011.13456},
  year={2020}
}

@inproceedings{choi2018stargan,
  title={Stargan: Unified generative adversarial networks for multi-domain image-to-image translation},
  author={Choi, Yunjey and Choi, Minje and Kim, Munyoung and Ha, Jung-Woo and Kim, Sunghun and Choo, Jaegul},
  booktitle={Proceedings of the IEEE conference on computer vision and pattern recognition},
  pages={8789--8797},
  year={2018}
}

@article{sun2017learning,
  title={Learning to write stylized chinese characters by reading a handful of examples},
  author={Sun, Danyang and Ren, Tongzheng and Li, Chongxun and Su, Hang and Zhu, Jun},
  journal={arXiv preprint arXiv:1712.06424},
  year={2017}
}

@inproceedings{wang2023cf,
  title={Cf-font: Content fusion for few-shot font generation},
  author={Wang, Chi and Zhou, Min and Ge, Tiezheng and Jiang, Yuning and Bao, Hujun and Xu, Weiwei},
  booktitle={Proceedings of the IEEE/CVF conference on computer vision and pattern recognition},
  pages={1858--1867},
  year={2023}
}

@article{chen2024if,
  title={IF-Font: Ideographic Description Sequence-Following Font Generation},
  author={Chen, Xinping and Ke, Xiao and Guo, Wenzhong},
  journal={Advances in Neural Information Processing Systems},
  volume={37},
  pages={14177--14199},
  year={2024}
}

@inproceedings{wen2021handwritten,
  title={Handwritten Chinese font generation with collaborative stroke refinement},
  author={Wen, Chuan and Pan, Yujie and Chang, Jie and Zhang, Ya and Chen, Siheng and Wang, Yanfeng and Han, Mei and Tian, Qi},
  booktitle={Proceedings of the IEEE/CVF winter conference on applications of computer vision},
  pages={3882--3891},
  year={2021}
}

@inproceedings{yao2024vq,
  title={Vq-font: Few-shot font generation with structure-aware enhancement and quantization},
  author={Yao, Mingshuai and Zhang, Yabo and Lin, Xianhui and Li, Xiaoming and Zuo, Wangmeng},
  booktitle={Proceedings of the AAAI Conference on Artificial Intelligence},
  volume={38},
  number={15},
  pages={16407--16415},
  year={2024}
}

@inproceedings{yang2024fontdiffuser,
  title={Fontdiffuser: One-shot font generation via denoising diffusion with multi-scale content aggregation and style contrastive learning},
  author={Yang, Zhenhua and Peng, Dezhi and Kong, Yuxin and Zhang, Yuyi and Yao, Cong and Jin, Lianwen},
  booktitle={Proceedings of the AAAI conference on artificial intelligence},
  volume={38},
  number={7},
  pages={6603--6611},
  year={2024}
}

@inproceedings{zhang2024dp,
  title={Dp-font: Chinese calligraphy font generation using diffusion model and physical information neural network},
  author={Zhang, Liguo and Zhu, Yalong and Benarab, Achref and Ma, Yusen and Dong, Yuxin and Sun, Jianguo},
  booktitle={Proceedings of the Thirty-Third International Joint Conference on Artificial Intelligence, IJCAI-24, K. Larson, Ed. International Joint Conferences on Artificial Intelligence Organization},
  volume={8},
  pages={7796--7804},
  year={2024}
}

@inproceedings{park2021few,
  title={Few-shot font generation with localized style representations and factorization},
  author={Park, Song and Chun, Sanghyuk and Cha, Junbum and Lee, Bado and Shim, Hyunjung},
  booktitle={Proceedings of the AAAI conference on artificial intelligence},
  volume={35},
  number={3},
  pages={2393--2402},
  year={2021}
}

@inproceedings{zhang2018separating,
  title={Separating style and content for generalized style transfer},
  author={Zhang, Yexun and Zhang, Ya and Cai, Wenbin},
  booktitle={Proceedings of the IEEE conference on computer vision and pattern recognition},
  pages={8447--8455},
  year={2018}
}

@inproceedings{zhu2017unpaired,
  title={Unpaired image-to-image translation using cycle-consistent adversarial networks},
  author={Zhu, Jun-Yan and Park, Taesung and Isola, Phillip and Efros, Alexei A},
  booktitle={Proceedings of the IEEE international conference on computer vision},
  pages={2223--2232},
  year={2017}
}

@article{pearton2000gan,
  title={GaN electronics},
  author={Pearton, Stephen J and Ren, Fan},
  journal={Advanced Materials},
  volume={12},
  number={21},
  pages={1571--1580},
  year={2000},
  publisher={Wiley Online Library}
}

@inproceedings{tang2022few,
  title={Few-shot font generation by learning fine-grained local styles},
  author={Tang, Licheng and Cai, Yiyang and Liu, Jiaming and Hong, Zhibin and Gong, Mingming and Fan, Minhu and Han, Junyu and Liu, Jingtuo and Ding, Errui and Wang, Jingdong},
  booktitle={Proceedings of the IEEE/CVF conference on computer vision and pattern recognition},
  pages={7895--7904},
  year={2022}
}

@article{sasaki2021unit,
  title={Unit-ddpm: Unpaired image translation with denoising diffusion probabilistic models},
  author={Sasaki, Hiroshi and Willcocks, Chris G and Breckon, Toby P},
  journal={arXiv preprint arXiv:2104.05358},
  year={2021}
}

@article{chicco2021coefficient,
  title={The coefficient of determination R-squared is more informative than SMAPE, MAE, MAPE, MSE and RMSE in regression analysis evaluation},
  author={Chicco, Davide and Warrens, Matthijs J and Jurman, Giuseppe},
  journal={Peerj computer science},
  volume={7},
  pages={e623},
  year={2021},
  publisher={PeerJ Inc.}
}

@article{gil2003efficient,
  title={Efficient dilation, erosion, opening, and closing algorithms},
  author={Gil, Joseph Yossi and Kimmel, Ron},
  journal={IEEE Transactions on Pattern Analysis and Machine Intelligence},
  volume={24},
  number={12},
  pages={1606--1617},
  year={2003},
  publisher={IEEE}
}

@inproceedings{han1995influence,
  title={The influence of the sigmoid function parameters on the speed of backpropagation learning},
  author={Han, Jun and Moraga, Claudio},
  booktitle={International workshop on artificial neural networks},
  pages={195--201},
  year={1995},
  organization={Springer}
}

@article{merris1994laplacian,
  title={Laplacian matrices of graphs: a survey},
  author={Merris, Russell},
  journal={Linear algebra and its applications},
  volume={197},
  pages={143--176},
  year={1994},
  publisher={Elsevier}
}

@inproceedings{fu2024generate,
  title={Generate like experts: Multi-stage font generation by incorporating font transfer process into diffusion models},
  author={Fu, Bin and Yu, Fanghua and Liu, Anran and Wang, Zixuan and Wen, Jie and He, Junjun and Qiao, Yu},
  booktitle={Proceedings of the IEEE/CVF conference on computer vision and pattern recognition},
  pages={6892--6901},
  year={2024}
}

@inproceedings{wang2025mx,
  title={MX-Font++: Mixture of Heterogeneous Aggregation Experts for Few-shot Font Generation},
  author={Wang, Weihang and Sun, Duolin and Zhang, Jielei and Gao, Longwen},
  booktitle={ICASSP 2025-2025 IEEE International Conference on Acoustics, Speech and Signal Processing (ICASSP)},
  pages={1--5},
  year={2025},
  organization={IEEE}
}

@inproceedings{xie2021dg,
  title={Dg-font: Deformable generative networks for unsupervised font generation},
  author={Xie, Yangchen and Chen, Xinyuan and Sun, Li and Lu, Yue},
  booktitle={Proceedings of the IEEE/CVF conference on computer vision and pattern recognition},
  pages={5130--5140},
  year={2021}
}

@inproceedings{liu2022xmp,
  title={Xmp-font: Self-supervised cross-modality pre-training for few-shot font generation},
  author={Liu, Wei and Liu, Fangyue and Ding, Fei and He, Qian and Yi, Zili},
  booktitle={Proceedings of the IEEE/CVF conference on computer vision and pattern recognition},
  pages={7905--7914},
  year={2022}
}

@inproceedings{liu2019few,
  title={Few-shot unsupervised image-to-image translation},
  author={Liu, Ming-Yu and Huang, Xun and Mallya, Arun and Karras, Tero and Aila, Timo and Lehtinen, Jaakko and Kautz, Jan},
  booktitle={Proceedings of the IEEE/CVF international conference on computer vision},
  pages={10551--10560},
  year={2019}
}

@article{wu2020calligan,
  title={Calligan: Style and structure-aware chinese calligraphy character generator},
  author={Wu, Shan-Jean and Yang, Chih-Yuan and Hsu, Jane Yung-jen},
  journal={arXiv preprint arXiv:2005.12500},
  year={2020}
}

@misc{foundertype,
  author = {{Beijing Founder Electronics Co., Ltd.}},
  title = {Foundertype Font Library},
  year = {2024},
  howpublished = {\url{https://www.foundertype.com/}},
  note = {Accessed: 2025-12-20}
}

\end{document}